\pdfoutput=1

\documentclass[11pt]{article}
\usepackage{authblk}

\usepackage{acl}

\usepackage{times}
\usepackage{latexsym}
\usepackage{booktabs}
\usepackage{array,multirow,graphicx}

\usepackage{lipsum}
\usepackage{graphicx}
\usepackage{amsmath}
\usepackage{bbm}
\usepackage{hyperref}
\usepackage{float}

\usepackage[T1]{fontenc}

\usepackage[utf8]{inputenc}

\usepackage{microtype}

\DeclareUnicodeCharacter{0307}{}

\title{On the Benefits of Fine-Grained Loss Truncation: A Case Study on Factuality in Summarization}

\author[1]{\bf Lorenzo Flores}
\author[1,2]{\bf Arman Cohan}

\affil[1]{Yale University\quad $^2$Allen Institute for AI}

\begin{document}
\maketitle
\begin{abstract}

Text summarization and simplification are among the most widely used applications of AI. However, models developed for such tasks are often prone to hallucination, which can result from training on unaligned data. 
One efficient approach to address this issue is Loss Truncation (LT) \citep{kang-hashimoto-2020-improved}, an approach to modify the standard log
loss to adaptively remove noisy examples during training.
However, we find that LT alone yields a considerable number of hallucinated entities on various datasets. We study the behavior of the underlying losses between factual and non-factual examples, to understand and refine the performance of LT. We demonstrate that LT's performance is limited when the underlying assumption that noisy targets have higher NLL loss is not satisfied, and find that word-level NLL among \textit{entities} provides better signal for distinguishing factuality. We then leverage this to propose a fine-grained NLL loss and fine-grained data cleaning strategies, and observe improvements in hallucination reduction across some datasets. Our work is available at \url{https://https://github.com/yale-nlp/fine-grained-lt}.

\end{abstract}

\section{Introduction}

Text summarization and simplification stand among the most popular applications in natural language generation (NLG). Yet, the models developed for these tasks suffer from generating inaccurate or unsupported information (AKA hallucinations) \citep{cao-etal-2022-hallucinated, zhao-etal-2020-reducing, maynez-etal-2020-faithfulness, tang-etal-2023-understanding}. This issue poses a significant risk of negative real-world consequences and can impede the wider adoption of these models.

To mitigate hallucinations, previous work studied aspects of training \citep{choubey-etal-2023-cape}, decoding \citep{van-der-poel-etal-2022-mutual, king-etal-2022-dont, sridhar-2022-improved}, or post-processing \citep{chen-etal-2021-improving}. In this paper, however, we focus on another crucial source of hallucination: the training data. 
When training data is misaligned (i.e. ground truth targets contain information unsupported by the input), models learn these patterns and hallucinate \citep{ji-etal-2023-survey, dziri-etal-2022-origin}. This can stem from data collection errors, or scraping web-based data \citep{ji-etal-2023-survey}. While there have been efforts to rectify the misaligned examples \citep{goyal-durrett-2021-annotating, ladhak-etal-2023-contrastive, zhou-etal-2021-detecting, adams-etal-2022-learning, filippova-2020-controlled, wan-bansal-2022-factpegasus}, these solutions typically involve either rewriting ground truth targets or training additional models to identify hallucinations, which presents its own set of challenges.

To sidestep such challenges, other methods automatically detect and remove noisy examples from training data. One simple and efficient approach is \textit{Loss Truncation (LT)} \citep{kang-hashimoto-2020-improved}, which filters out noisy examples based on the observation that they have higher negative log-likelihood (NLL) loss. This enables an easy-to-adapt and highly efficient training procedure: if NLL loss is high (e.g. >80th quantile of observed losses), the loss is \emph{not} backpropagated. This mehod is widely adopted in summarization \citep{guo-etal-2021-resilient, ladhak-etal-2022-faithful, cao-etal-2022-learning, goyal-etal-2022-training, hewitt-etal-2022-truncation}, however, applying LT to five datasets, we find that models continue to produce a significant amount of hallucinated content.

In this paper, we study the behavior of NLL at the original (i.e. sentence) and fine-grained level (i.e. token) to understand and refine the performance of LT. At the time of writing, the paper is the first to analyze LT on text simplification datasets like Cochrane, MedEasi, and ASSET; moreover, it analyzes the performance of LT from the perspective of factuality, and delves deeper into training dynamics at the token and entity level. Ultimately, our work aims to contribute a better understanding of the underlying dynamics of LT, offering insights that could inform its future applications, particularly in efforts to reduce hallucination.

We make the following contributions: (1) We demonstrate that LT's performance is hindered when the underlying assumption that noisy targets have higher NLL loss is not satisfied, (2) we find that word-level NLL among \textit{entities} provides better signal for distinguishing factuality, and (3) we use this to propose a fine-grained NLL loss which reduces entity-level hallucination on some datasets (-22\% on Cochrane, -7.2\% on ASSET), and fine-grained data cleaning strategies which achieve up to 26.8\% hallucination reduction (CNN-DM), highlighting the potential of this approach.

\section{Methodology}

\paragraph{Loss Truncation} 
Loss Truncation \citep{kang-hashimoto-2020-improved, goyal-etal-2022-training, cao-etal-2022-learning} is an efficient method for improving language generation by modifying the standard log
loss to adaptively disregard examples with high loss, reducing potential hallucinations. 
It continuously updates a list of example-level NLL losses, and zeros out losses above a set quantile.\footnote{We adapt the implementation by \citet{kang-hashimoto-2020-improved} into a plug-and-play library for training losses \url{https://github.com/ljyflores/loss-library}}

Formally, Loss Truncation defines the loss as \begin{align*}
\text{NLL} &= - \sum_{t=1}^{|y|} y_t \text{log} (\hat{y}_t) \\
\mathcal{L}_\text{LT} &= \text{NLL} \cdot \mathbbm{1} [ \text{NLL} < \text{cutoff} ] 
\end{align*}

\paragraph{Datasets} We study two popular conditional NLG tasks, summarization and simplification, where data sources can be noisy, and hallucinations remain an issue. We select five datasets representing a variety of domains: \textbf{Cochrane} \citep{devaraj-etal-2021-paragraph}: Medical abstracts from Cochrane Database of Systematic Reviews and expert-written summaries (4,459 pairs), \textbf{MedEasi} \citep{basu-etal-2023-medeasi}: Sentences from Merck Manuals \citep{cao-etal-2020-expertise} and SimpWiki \citep{VanDenBercken2019SimpWiki} and annotated simplifications (1,697 pairs), \textbf{ASSET} \citep{alva-manchego-etal-2020-asset}: Sentences from TurkCorpus dataset \citep{xu-etal-2016-optimizing} and simplified versions by 10 annotators (23,590 pairs), \textbf{CNN/DailyMail} \cite{nallapati-etal-2016-abstractive}: Articles and their highlight summaries from CNN and DailyMail (311,971 pairs), \textbf{XSum} \cite{narayan-etal-2018-dont}: BBC news articles and their corresponding one-line summaries (226,711 pairs).

\paragraph{Models} We use BART-Large-XSUM \citep{lewis-etal-2020-bart} as the base model and fine-tune it on each dataset. 
We select BART as it is the base model studied for LT in prior research \citep{lu-etal-2023-napss, devaraj-etal-2021-paragraph,martin-etal-2022-muss,cao-etal-2022-learning}, and allows us to specifically study its effects. We further experiment with FlanT5 as a stronger base model \citep{chung2022scaling} with LT for comparison, and find that it yields similar or better performance (Appendix \ref{Appendix:FlanT5_Results}). 

\paragraph{Finetuning} We first do standard seq2seq training on a BART-XSum model as a baseline (BART-XSum FT). We then finetune a separate BART-XSum model with the original LT, and another with fine LT. Additionally, we finetune two other models with seq2seq training, but on two cleaned versions of the original datasets, labelled as Drop Sentence and Drop Example (See Appendix \ref{Appendix:LT_Hyperparams} for details on implementation). We also provide evaluation results from prior work for comparison.

\paragraph{Entity-Based Hallucination} We identify a summary to have ``hallucination'' with respect to a source if the summary contains an unsupported entity (i.e. an entity which is not found in the reference). 
To determine this, we first use NER models to find the entities in both the source text and the summary\footnote{We use \texttt{en\_core\_web\_lg} and \texttt{en\_core\_sci\_lg} NER \citep{spacy-2017, neumann-etal-2019-scispacy}}. Note that we filter the entity types down to dates, numbers, proper nouns, and specific medical conditions (for MedEasi and Cochrane) to ensure that synonyms are not involved in this process. If any of the entities in the summary are not present in the source, we say the summary contains a hallucination. While this can be perceived as a strong assumption, our pilot manual observations indicated that unsupported entities of these type are often indeed hallucinations, hence this heuristic works well in practice (See Appendix \ref{Appendix:Example_Hallucinations_Eval}).

\paragraph{Metrics} We propose a simple definition as our metric of factuality, Hallucination Rate (HR): the \% of outputs containing a hallucination. We also evaluate faithfulness using the question-answering based QuestEval \citep{scialom-etal-2021-questeval} metric. In addition, we evaluate overall fluency with SARI \citep{xu-etal-2016-optimizing}, an edit-based text simplification metric, and ROUGE-LSum \citep{lin-2004-rouge}, computed using EASSE to align our work with previous methods \citep{alva-manchego-etal-2019-easse}.

\paragraph{Experimental setup} We first finetune a BART Sum model using LT \citep{kang-hashimoto-2020-improved}, and study its ability to reduce hallucination (Table \ref{Table:Results}). Although LT reduces hallucination compared to previous SOTA models, we find that many outputs still contain hallucinations. This leads us to study the validity of the assumptions underlying LT, which may explain its relatively poor performance. In particular, we study the assumption that noisy examples have higher NLL loss, by comparing the training losses of clean and noisy examples (Table \ref{Table:NLL_by_Epoch}). This analysis reveals that the difference in NLL loss between clean and noisy examples is marginal, thus explaining LT’s inability to distinguish the two. We then probe the losses at the token level to determine if losses on certain tokens are better indicators of noise (Tables \ref{Table:NLL_by_Token}, \ref{Table:F_vs_NF_Entity_Comparison}). Our findings lead us to propose a ``fine-grained LT'' and heuristic data cleaning strategies (Table \ref{Table:Results}).

\section{Findings}

\paragraph{Noise in summarization can come from adding unsupported information in the reference}

Our experiments are motivated by the observation that some target outputs contained unsupported information (see Appendix \ref{Appendix:Noisy_Examples}). E.g., some references in Cochrane had the phrase ``\textit{The evidence is current to [date]}'', although the date was not mentioned in the input. Upon finetuning, models learn to reproduce this pattern with incorrect dates (Appendix \ref{Appendix:Example_Output}). This illustrates how datasets can be noisy, especially when there is irrelevant information added to the reference summary \citep{ji-etal-2023-survey}. Hence, we limit our definition of \textit{noisy} and \textit{hallucination} as containing unsupported data; we call references containing unsupported entities as noisy. 

\begin{table}[h]
\centering
\footnotesize
\begin{tabular}{@{}llrrrrrr@{}}
\toprule
\textbf{Data} & \textbf{Model} & \textbf{HR} $\downarrow$ & \textbf{SR} $\uparrow$ & \textbf{RL} $\uparrow$ & \textbf{QE} $\uparrow$ \\
\midrule
\parbox[t]{4mm}{\multirow{7}{*}{\rotatebox[origin=c]{90}{Cochrane}}} \parbox[t]{4mm}{\multirow{4}{*}{\rotatebox[origin=c]{90}{Previous}}} & BART FT & 69.3\% & 35.6 & 44.7 & \textbf{0.546} \\
 & UL \citeyearpar{devaraj-etal-2021-paragraph} & 69.6\% & \textbf{40.0} & 39.2 & 0.495\\
 & NA \citeyearpar{lu-etal-2023-napss} & 73.8\% & 32.9 & \textbf{45.4} & 0.523 \\
 & LT \citeyearpar{kang-hashimoto-2020-improved} & 42.7\% & 36.2 & 37.6 & 0.517 \\
 \cmidrule{2-6}
\parbox[t]{4mm}{\multirow{3}{*}{\rotatebox[origin=c]{90}{}}} \parbox[t]{4mm}{\multirow{3}{*}{\rotatebox[origin=c]{90}{Ours}}} & LT Fine & \textbf{20.6\%} & 36.1 & 21.8 & 0.446 \\
 & Drop Sent & 42.1\% & 38.6 & 33.7 & 0.470 \\
 & Drop Ex & 37.1\% & 38.5 & 31.9 & 0.482 \\
\midrule
\parbox[t]{4mm}{\multirow{7}{*}{\rotatebox[origin=c]{90}{MedEasi}}} \parbox[t]{4mm}{\multirow{4}{*}{\rotatebox[origin=c]{90}{Previous}}} & BART FT & 35.7\% & \textbf{40.5} & 45.7 & 0.588 \\
& UL \citeyearpar{devaraj-etal-2021-paragraph}* & 13.7\% & 35.3 & \textbf{47.9} & 0.653 \\
& NA \citeyearpar{lu-etal-2023-napss}* & 42.3\% & 34.0 & 24.3 & 0.418 \\
& LT \citeyearpar{kang-hashimoto-2020-improved} & \textbf{4.6\%} & 32.6 & 47.3 & \textbf{0.656} \\
 \cmidrule{2-6}
\parbox[t]{4mm}{\multirow{3}{*}{\rotatebox[origin=c]{90}{}}} \parbox[t]{4mm}{\multirow{3}{*}{\rotatebox[origin=c]{90}{Ours}}} & LT Fine & 7.0\% & 37.9 & 45.1 & 0.628 \\
& Drop Sent & 7.0\% & 31.8 & 47.5 & 0.622 \\
& Drop Ex & 9.7\% & 38.9 & 44.4 & 0.651 \\
\midrule
\parbox[t]{4mm}{\multirow{7}{*}{\rotatebox[origin=c]{90}{ASSET}}} \parbox[t]{4mm}{\multirow{4}{*}{\rotatebox[origin=c]{90}{Previous}}} & BART FT & 17.0\% & 38.9 & \textbf{86.0} & 0.706 \\
 & M - \citeyearpar{martin-etal-2022-muss} & 23.4\% & 43.6 & 81.4 & 0.706 \\
 & M + \citeyearpar{martin-etal-2022-muss} & 31.5\% & \textbf{44.1} & 79.4 & 0.693 \\
 & LT \citeyearpar{kang-hashimoto-2020-improved} & 14.2\% & 36.7 & 77.7 & 0.659 \\
 \cmidrule{2-6}
 \parbox[t]{4mm}{\multirow{3}{*}{\rotatebox[origin=c]{90}{}}} \parbox[t]{4mm}{\multirow{3}{*}{\rotatebox[origin=c]{90}{Ours}}} & LT Fine & \textbf{6.9\%} & 37.9 & 45.1 & \textbf{0.749} \\
 & Drop Sent & 12.8\% & 40.0 & 81.7 & 0.703 \\
 & Drop Ex & 22.3\% & 38.9 & 85.1 & 0.664 \\
\midrule
\parbox[t]{4mm}{\multirow{6}{*}{\rotatebox[origin=c]{90}{CNN}}} \parbox[t]{4mm}{\multirow{3}{*}{\rotatebox[origin=c]{90}{Previous}}} & BART FT & 68.1\% & 41.4 & 29.9 & 0.592 \\
 & BRIO \citeyearpar{liu-etal-2022-brio} & 51.9\% & \textbf{44.9} & \textbf{38.3} & \textbf{0.596} \\
 & LT \citeyearpar{kang-hashimoto-2020-improved} & 58.8\% & 40.7 & 29.0 & 0.586 \\
 \cmidrule{2-6}
 \parbox[t]{4mm}{\multirow{3}{*}{\rotatebox[origin=c]{90}{}}} \parbox[t]{4mm}{\multirow{3}{*}{\rotatebox[origin=c]{90}{Ours}}} & LT Fine & 61.3\% & 41.3 & 29.7 & 0.587 \\
 & Drop Sent & \textbf{32.0\%} & 42.3 & 34.5 & 0.591 \\
 & Drop Ex & 66.7\% & 41.8 & 30.4 & 0.589 \\
\midrule
\parbox[t]{4mm}{\multirow{6}{*}{\rotatebox[origin=c]{90}{XSum}}} \parbox[t]{4mm}{\multirow{3}{*}{\rotatebox[origin=c]{90}{Previous}}} & BART FT & 76.9\% & 47.6 & 35.2 & 0.465 \\
 & BRIO \citeyearpar{liu-etal-2022-brio} & 77.1\% & \textbf{50.6} & \textbf{40.1} & 0.465 \\
 & LT \citeyearpar{kang-hashimoto-2020-improved} & 72.6\% & 48.1 & 36.4 & \textbf{0.474} \\
 \cmidrule{2-6}
 \parbox[t]{4mm}{\multirow{3}{*}{\rotatebox[origin=c]{90}{}}} \parbox[t]{4mm}{\multirow{3}{*}{\rotatebox[origin=c]{90}{Ours}}} & LT Fine & 75.5\% & 47.1 & 34.5 & 0.465 \\
 & Drop Sent & 70.0\% & 47.2 & 34.9 & 0.466 \\
 & Drop Ex & \textbf{69.3\%} & 47.0 & 34.8 & 0.467 \\
\bottomrule
\end{tabular}
\caption{\label{Table:Results}
Hallucination Rate (HR), SARI (SR), and ROUGE-LSum (RL), computed using EASSE \citep{alva-manchego-etal-2019-easse}, and QuestEval with Reference (QE) \citep{scialom-etal-2021-questeval} computed using \href{https://github.com/ThomasScialom/QuestEval}{GitHub} implementation from one run; * We generate these results on MedEasi; FT: Finetuned, -: Not Mined, +: Mined}
\end{table}

\paragraph{LT reduces entity-level hallucination from noisy targets, but not completely}

\begin{table}[t]
\centering
\footnotesize
\begin{tabular}{lrrr}
\toprule
 \textbf{Dataset} & \textbf{NLL (-)} & \textbf{NLL (+)} & \textbf{$\Delta$} \\
\midrule
Cochrane & 8.438 & 9.077 & -0.639 \\
MedEasi & 11.114 & 11.173 & -0.058 \\
Asset & 11.197 & 11.196 & 0.002 \\
XSum & 19.187 & 19.190 & -0.003 \\
CNN & 10.813 & 10.830 & -0.017 \\
\midrule
Cochrane & 0.651 & 0.437 & 0.214* \\
MedEasi & 0.080 & 0.032 & 0.048* \\
Asset & 0.055 & 0.034 & 0.021* \\
XSum & 0.049 & 0.043 & 0.006* \\
CNN & 0.134 & 0.112 & 0.022* \\
\bottomrule
\end{tabular}
\caption{\label{Table:NLL_by_Epoch}
Average NLL Loss for Non-Factual (-) and Factual (+) Examples at Epoch 0 (top) and 1 (bottom), * Indicates the significant difference (One-Way Mann-Whitney Test, $\alpha=0.05$)}
\vspace{-8pt}
\end{table}

We finetune BART-XSum using LT \citep{kang-hashimoto-2020-improved}, expecting it to filter out noisy examples and thus mitigate hallucinations. While LT reduces hallucination compared to the baseline and previous work (See Table \ref{Table:Results}), a considerable proportion of examples still contain hallucinations. 

\paragraph{We hypothesize that LT's performance suffers because of the unmet assumption of higher NLL in noisy data} 

\begin{figure*}
\centering
  \includegraphics[width=14cm]{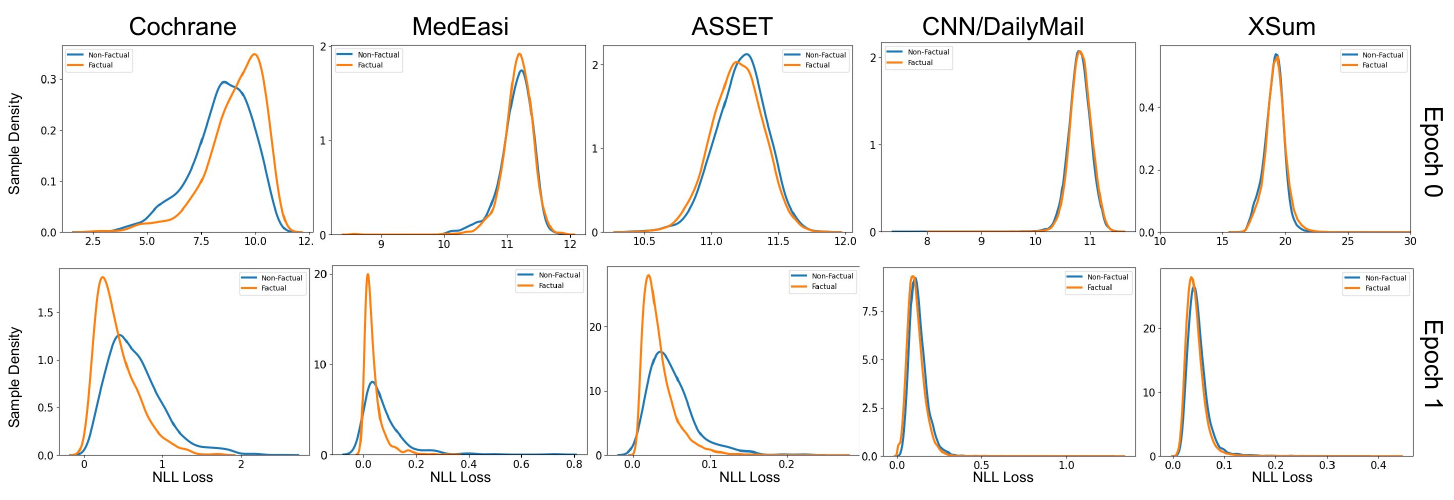}
  \caption{NLL distribution of \textcolor{orange}{factual} (\textcolor{orange}{Orange}) and \textcolor{blue}{non-factual} (\textcolor{blue}{Blue}) targets have no difference before finetuning, and a slight difference after one epoch of finetuning, with non-factual entities having slightly higher NLL \label{Figure:NLL_Plot_Ent}}
\end{figure*}

We seek to understand why LT is unable to weed out many hallucinated entities. We study the assumption that noisy examples have higher NLL loss, by comparing NLL loss of noisy and clean examples before finetuning and after 1 epoch of fine-tuning, when most models converge (See Appendix \ref{Appendix:Loss_Curves} for convergence information). Before finetuning, there is no significant difference between the NLL losses of factual (+) and non-factual (-) examples (Table \ref{Table:NLL_by_Epoch}, top). After one epoch of training, non-factual examples have higher NLL losses than factual sentences (Table \ref{Table:NLL_by_Epoch}, bottom). However, Figure \ref{Figure:NLL_Plot_Ent} shows that the difference in NLL is not large enough to cleanly distinguish factual (\textcolor{orange}{orange}) from non-factual (\textcolor{blue}{blue}) examples solely by their training losses. Recall that LT tries to separate the factual and non-factual training examples using a cutoff value. When the distributions of NLL losses of factual and non-factual examples overlap significantly like in Figure \ref{Figure:NLL_Plot_Ent}, it will be difficult to find one cutoff which separates the two distributions. This explains LT’s limited performance: noisy examples’ NLL is not always higher than non-noisy examples, which prevents LT from identifying and removing noisy examples. 

\paragraph{Word-level NLL may better distinguish between factual vs non-factual examples} Because NLL at the example level is insufficient to distinguish factual from non-factual examples, we study token-level NLL losses to see if particular tokens may provide better distinction. 

We make two observations: First, we find that within non-factual sentences, non-factual entities (\textbf{NLL (-)}) generally have higher NLL than factual entities (\textbf{NLL (+)}) (Table \ref{Table:NLL_by_Token}). Moreover, the magnitude of the difference in NLL (\textbf{$\Delta$}) is larger at the entity level than the sentence level (i.e. compared to the $\Delta$ column in Table \ref{Table:NLL_by_Epoch}).

We then check how the NLL of entity tokens in non-factual sentences compare to those in factual sentences (Table \ref{Table:F_vs_NF_Entity_Comparison}). Again, we find that non-factual entities have higher NLL than factual entities. This suggests that the NLL of entity tokens may provide better signal as to whether an example is noisy.

\begin{table}[t]
\centering
\footnotesize
\begin{tabular}{lrrrr}
\toprule
 \textbf{Dataset} & \textbf{NLL (0)} & \textbf{NLL (-)} & \textbf{NLL (+)} & \textbf{$\Delta$} \\
\midrule
Cochrane & 8.621 & 2.445 & 0.601 & 1.844* \\
MedEasi & 11.161 & 2.231 & 0.772 & 1.458* \\
Asset & 11.192 & 2.550 & 0.664 & 1.886* \\
XSum & 19.045 & 1.865 & 1.934 & -0.068* \\
CNN & 10.852 & 2.910 & 2.083 & 0.827* \\
\midrule
Cochrane & 0.669 & 1.592 & 0.331 & 1.261* \\
MedEasi & 0.078 & 2.070 & 0.443 & 1.626* \\
Asset & 0.051 & 3.392 & 0.300 & 3.092* \\
XSum & 0.048 & 0.946 & 1.354 & -0.409 \\
CNN & 0.128 & 1.842 & 1.447 & 0.395* \\
\bottomrule
\end{tabular}
\caption{\label{Table:NLL_by_Token}
Average NLL Loss for Non-Entity (0), Non-Factual Entity (-) and Factual Entity (+) Tokens at Epoch 0 (top) and 1 (bottom), * Indicates the significant difference (One-Way Mann-Whitney Test, $\alpha=0.05$)}
\vspace{-8pt}
\end{table}

Second, an example’s NLL is largely influenced by non-entity tokens. This is shown by the fact that non-entity NLL values closely mirror the sentence-level NLLs (Table \ref{Table:NLL_by_Epoch}, NLL (-)). Intuitively it makes sense: there are more non-entities than entities, so they have a larger impact on sentence-level NLL. 

However, these tokens obscure the signal between factual and non-factual entities. This suggests that using only the entity tokens to compute the LT cutoff may provide LT with more signal to distinguish factual vs non-factual examples.

\paragraph{We propose a fine-grained LT, which reduces hallucination on moderately noisy datasets}

We first propose a fine-grained LT method: instead of using sentence-level NLL in LT, we sum the NLL \textit{only} for entity tokens. This leverages the fact that entity tokens provide better signal for whether an example is noisy. Formally, fine-grained LT is given by \begin{align*}
\text{score} &= \sum_{t=1}^{|y|} \mathbbm{1} [ y_t \in \text{entities} ] \cdot y_t \text{log} (\hat{y}_t) \\
\mathcal{L}_\text{LT-Fine} &= \text{NLL} \cdot \mathbbm{1} [ \text{score} < \text{cutoff} ] 
\end{align*}

where $\mathbbm{1} [ y_t \in \text{entities} ]$ is scored by Spacy / SciSpacy NER models \citep{spacy-2017, neumann-etal-2019-scispacy} and $\hat{y_t} = p(y_t|y_{<t}, X)$.

Fine-grained LT reduces HR on Cochrane (-22\%) and ASSET (-7.2\%) compared to original LT (Table \ref{Table:Results}). However, its performance is not as competitive on MedEasi, CNN, and XSum. 

We hypothesize this is because the three datasets are web-scraped and noisier, unlike Cochrane and ASSET which are human annotated. We confirm this by computing HR using the datasets' source text and its own labels, using on 100 labels in each of the datasets' test sets. Indeed, labels from the three web-scraped dataset contained more hallucinated entities than the human annotated ones (Table \ref{Table:Gold_Label_HR}, Appendix \ref{Appendix:Noisy_Examples} for examples). This can stem from misalignment between the source and label: For example, a news dataset may use the body of a news article as the source text, and the headline as the reference summary. There may be names mentioned in the headline which do not appear in the news article. Following our definition, these names would be considered as hallucinations. Therefore, we suspect these datasets require a more aggressive strategy to eliminate such noise.

\paragraph{For noisier datasets, we propose fine-grained data cleaning strategies to reduce hallucination}

To this end, we directly clean the dataset, filtering out noisy targets. We identify all unsupported entities in a target (i.e. the entity is not in the input); then we either (1) drop \textit{only} the sentence containing the entity (Drop Sentence), or (2) drop the entire example (Drop Example) (Table \ref{Table:Data_Clean_Stats}). 

\begin{table}[]
\centering
\footnotesize
\begin{tabular}{@{}lrrr@{}}
\toprule
\textbf{Dataset} & \textbf{Original} & \textbf{Drop Sentence} & \textbf{Drop Example} \\
\midrule
Cochrane & 3568 & 3479 & 245 \\
MedEasi & 1397 & 907 & 857 \\
ASSET & 20000 & 18690 & 18229 \\
CNN & 287113 & 285160 & 187465 \\
XSum & 204045 & 110754 & 110745 \\
\bottomrule
\end{tabular}
\caption{\label{Table:Data_Clean_Stats} Number of training examples from data cleaning methods; Drop Sentence results in minor reductions whereas Drop Example results in larger reductions}
\end{table}
\vspace{-3pt}

Table \ref{Table:Results} shows that at least one of the strategies results in lower hallucination rate for CNN (-26.8\%, Drop Sentence) and XSum (-3.3\%, Drop Example), and competitive performance with SOTA for the MedEasi dataset. In addition to reducing hallucination, we note that our methods achieve competitive performance on SARI and QuestEval (Table \ref{Table:Results}), demonstrating that our methods can reduce hallucination without significantly affecting models' overall fluency and faithfulness. Except for MedEasi dataset, our results show strong improvements over the baseline methods, suggesting the potential of the fine-grained LT and fine-grained data cleaning in reducing hallucinations.

\section{Conclusion}

We analyzed the effect of loss truncation (LT) on improving factuality in text summarization. We found that LT struggles to reduce entity-level hallucination when the underlying assumption that non-factual sentences have higher NLL than factual sentences is not met. To this end, we explore a token-level loss truncation (i.e. fine-grained LT) and simple entity-level dataset cleaning strategies, which reduce the prevalence of hallucination across various summarization and simplification datasets.

Future work may explore other signals for noise in training data. Moreover, future work can explore contradictory information (i.e. targets with similar topics as input but different meaning). This requires the use of natural language inference (NLI), which we qualitatively find is difficult in practice using off-the-shelf NLI models \citep{wu-etal-2022-generating} or GPT \citep{liu-2023-geval}, as we observe they are currently unable to detect contradictory or unsupported information in some cases. Ultimately, reducing such hallucinations is key to improving the overall performance of summarization models.

\section*{Limitations}
One limitation of our paper is that we limit the definition of hallucination to the addition of unsupported entities, while the detection of contradictory or omitted information are equally important to detect. A key challenge with such definitions of hallucination is that they require human annotations or good models to identify targets in the dataset which contain contradictory or omitted information. We previously experimented with using GPT-4 following the GPT-Eval framework \citep{liu-2023-geval}, but found that GPT was sometimes unable to detect unsupported information. For example, GPT was unable to identify that the date in the Cochrane dataset targets were unsupported.

Another limitation is that loss truncation at the token level does not always achieve the best results. While it reduced entity-level hallucination for Cochrane and ASSET compared to other methods, it fails to achieve substantial improvements on MedEasi, CNN, and XSum. Overall, the paper aims to show that the method has potential in some cases, but future work can explore other ways to improve its performance.

Finally, it should be noted that our work has been tested on a limited number of general domain summarization datasets; hence more work can explore a wider set of datasets in various niches, to examine if larger patterns across datasets impact the performance of loss truncation.

\paragraph{Risks} It should be noted that even data cleaning and LT (both original and fine-grained) does not fully eliminate entity-level hallucination. Moreover, we have not studied other types of hallucination in this work (e.g., discussion of supported entities in a wrong way). Therefore, caution should be employed when deploying our approaches in practice.

\bibliography{anthology,custom}

\appendix

\section{Dataset Information}
\label{Appendix:Data_Clean_Stats}

\paragraph{Licenses} The Cochrane dataset uses the C.C. BY 4.0 License; MedEasi and XSum use the MIT License; ASSET uses the CC BY-NC 4.0 License, and CNN/DailyMail uses the Apache 2.0 License.

\section{Dataset Noisiness Investigation}

We report the hallucination rates for the noise in the labels of 5 datasets, computed using the first 100 examples of their test sets. Note that for Cochrane, we manually removed common words which medical NER models thought were entities (e.g. ``disease'', ``operation'').

\begin{table}[h!]
\centering
\footnotesize
\begin{tabular}{rl}
\toprule
\textbf{Dataset} & \textbf{HR} $\downarrow $\\
\midrule
Cochrane & 68/100 \\
ASSET & 14/100 \\ 
MedEasi & 80/100 \\
CNN & 74/100 \\
XSum & 83/100 \\
\bottomrule
\end{tabular}
\caption{\label{Table:Gold_Label_HR} Noisiness of datasets measured using 100 examples' hallucination rate (HR)}
\end{table}
\vspace{-3pt}

\section{Training Details}
\label{Appendix:LT_Hyperparams}

\paragraph{Implementation Details} We run our experiments on 1 NVIDIA RTX 6000 GPU. Finetuning each model on Cochrane, MedEasi, and ASSET, for base, original and fine-grained LT, and with cleaned datasets, takes roughly 40 minutes, whereas CNN/DailyMail and XSum take 4 hours.

\paragraph{Finetuning} All models use 1 epoch, a learning rate of 5e-5, Adam epsilon of 1e-8, and batch size of 1 for Cochrane/MedEasi and 64 for ASSET, XSum, CNN/DailyMail). 

\paragraph{Loss Truncation (Original)} All datasets are trained using a 80\% truncate rate, with a cutoff recomputed every 1000 examples. 

\paragraph{Loss Truncation (Fine-Grained)} Cochrane and MedEasi use an 80\% truncate rate, whereas ASSET, XSum, and CNN/DailyMail use a 40\% truncate rate, all recomputing every 500 examples.

\section{Training Loss Curves}
\label{Appendix:Loss_Curves}

\begin{figure}[h]
\centering
  \includegraphics[width=\columnwidth]{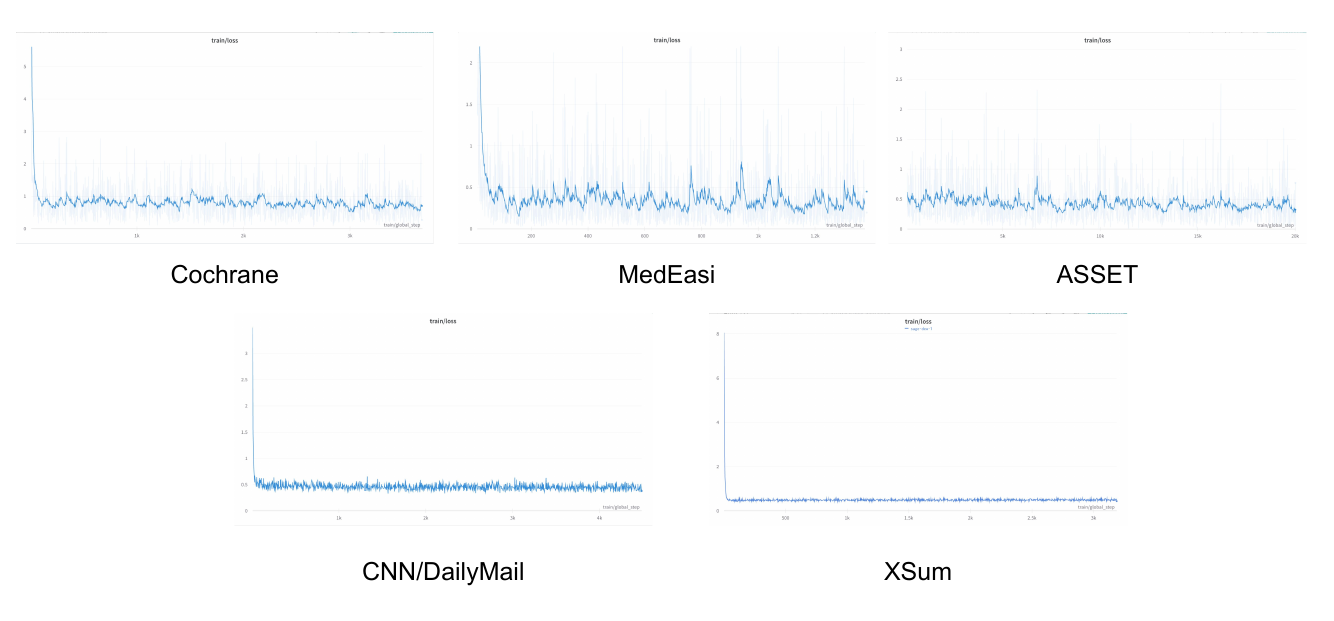}
  \caption{\label{Figure:Loss_Curves} Loss curves from finetuned BART-XSum; 0.8 smoothing used in top row}
\end{figure}

We plot loss curves generated from finetuning BART-XSum in Figure \ref{Figure:Loss_Curves} throughout one epoch which demonstrates convergence across datasets.

\section{NLL of Factual/Non-Factual Tokens}
\label{Appendix:NLL_Across_Sents}

We compare the NLL of factual and non-factual tokens in factual and non-factual sentences in Table \ref{Table:F_vs_NF_Entity_Comparison}. 
This demonstrates that non-factual tokens have higher NLL than factual tokens, regardless of which sentences the tokens appear in.

\begin{table}[]
\centering
\footnotesize
\begin{tabular}{lrrr}
\toprule
 \textbf{Dataset} & \textbf{NLL (+, NF)} & \textbf{NLL (+, F)}  & \textbf{NLL (-)}\\
\midrule
Cochrane & 0.601 & 0.522 & 2.445 \\
MedEasi & 0.772 & 0.510 & 2.231 \\
Asset & 0.664 & 0.752 & 2.550 \\
XSum & 1.934 & 2.579 & 1.865 \\
CNN & 2.083 & 2.199 & 2.910 \\
\midrule
Cochrane & 0.331 & 0.265 & 1.592 \\
MedEasi & 0.443 & 0.228 & 2.070 \\
Asset & 0.300 & 0.825 & 3.392 \\
XSum & 1.354 & 1.776 & 0.946 \\
CNN & 1.447 & 1.488 & 1.842 \\
\bottomrule
\end{tabular}
\caption{\label{Table:F_vs_NF_Entity_Comparison} Token-Level NLL Loss for Factual Entities in Non-Factual Targets (+, NF) and Factual Targets (+, F), and Non-Factual Entities in Non-Factual Targets (-)}
\end{table}

\section{Results on Flan-T5}
\label{Appendix:FlanT5_Results}

We report the details of finetuning the standard loss truncation \citep{kang-hashimoto-2020-improved} using Flan-T5 \citep{chung2022scaling} in Table \ref{Table:FlanT5_Results}.

\begin{table}[H]
\centering
\footnotesize
\begin{tabular}{lrrrrr}
\toprule
\textbf{Data} & \textbf{HR (Entity)} $\downarrow$ & \textbf{SARI} $\uparrow$ & \textbf{RL} $\uparrow$ \\
\midrule
Cochrane & 190/480 (39.6\%) & 33.720 & 37.163 \\
MedEasi & 14/300 (46.7\%) & 24.405 & 48.248 \\
ASSET & 19/359 (5.3\%) & 35.003 & 91.116 \\
CNN & 2948/11490 (25.7\%) & 41.486 & 32.133 \\
XSum & 6897/11334 (60.9\%) & 43.767 & 29.130 \\
\bottomrule
\end{tabular}
\caption{\label{Table:FlanT5_Results} Finetuning Flan-T5 \citep{chung2022scaling} with Loss Truncation results in even better performance than BART, demonstrating opportunity for further progress}
\end{table}
\vspace{-8pt}

\section{Examples of Noisy Targets}
\label{Appendix:Noisy_Examples}

See Table \ref{Table:Example_Noisy_Targets} for noisy targets from various datasets.

\begin{table*}[ht]
\centering
\begin{tabular}{p{1.5cm} p{6.5cm} p{6.5cm}}
\toprule
\textbf{Dataset} & \textbf{Input} & \textbf{Target} \\
\midrule
MedEasi &  Baker cysts may form and rupture. & Cysts may develop and rupture \textbf{behind the knees, suddenly increasing the pain}. \\
\cmidrule{2-3}
 & Sullivan apparently had no idea who McCartney was. & Sullivan thought that \textbf{his illness was because of ulcers}. \\
\cmidrule{2-3}
 & The linear combination of atomic orbitals or LCAO approximation for molecular orbitals was introduced in 1929 by Sir John Lennard-Jones. & The \textbf{LCMO (Linear combination of atomic orbitals molecular orbital)} method gives a rough but good description of the MOs \\
\cmidrule{1-3}
Cochrane & We included six trials, involving a total of 636 women with a twin or triplet pregnancy (total of 1298 babies). We assessed all of the included trials as having a low risk of bias for random sequence generation. ... There is a need for large-scale, multicenter randomised controlled trials to evaluate the benefits, adverse effects and costs of bed rest before definitive conclusions can be drawn. & \textbf{We searched for evidence on 30 May 2016.} We identified six randomised controlled trials involving a total of 636 women and 1298 babies ... We did not find sufficient evidence to support or refute bed rest for women with a multiple pregnancy as a way of preventing preterm birth and other pregnancy complications. \\ 
\cmidrule{2-3}
 & This update identified one additional study for inclusion, adding data for 2305 participants. This addition more than doubled the overall number of patients eligible for the review. ... Also, there were no data from RCTs on the utility of non-vitamin K antagonist oral anticoagulants compared to antiplatelet agents in heart failure with sinus rhythm. & This is an update of an earlier review. \textbf{The evidence is current to September 2015.} We only identified one new study with 2305 participants. ... It is unlikely that further studies will change these conclusions unless new, more effective and safe drugs become available. \\
\cmidrule{2-3}
XSum & The full cost of damage in Newton Stewart, one of the areas worst affected, is still being assessed. Repair work is ongoing in Hawick and many roads in Peeblesshire remain badly affected by standing water.... Have you been affected by flooding in Dumfries and Galloway or the Borders? & Clean-up operations are continuing across the Scottish Borders and Dumfries and Galloway after flooding caused by \textbf{Storm Frank}. \\
 & A fire alarm went off at the Holiday Inn in Hope Street at about 04:20 BST on Saturday and guests were asked to leave the hotel. As they gathered outside they saw the two buses, parked side-by-side in the car park, engulfed by flames.... While the exact cause is still under investigation, it is thought that the fire was started deliberately. & Two tourist buses have been destroyed by fire in a suspected arson attack in \textbf{Belfast city centre.} \\
\bottomrule
\end{tabular}
\caption{\label{Table:Example_Noisy_Targets} Examples of noisy targets from datasets, with the unsupported entities presented in \textbf{bold}}
\end{table*}

\section{Example Output}

See Table \ref{Table:Example_Output_1} for a comparison of outputs of various models. Loss truncation and the example-level data cleaning are the only methods which correctly avoid generating a hallucinated date.

\label{Appendix:Example_Output}

\begin{table*}[p]
\centering
\begin{tabular}{p{3cm} p{11.5cm}}
\toprule
\textbf{Model} & \textbf{Example} \\
\midrule
BART \citep{lewis-etal-2020-bart} & We found one small randomised controlled trial (with data from 34 women) that compared transcervical amnioinfusion with no treatment. The trial was considered to be at a high risk of bias overall… \textbf{The evidence is up-to-date as of February 2019.}\\
\cmidrule{2-2}
Both-UL \citep{devaraj-etal-2021-paragraph} & We found one small randomised controlled trial with 34 women…  \textbf{The evidence is current up-to-date to September 2016}… The numbers included were too small for meaningful assessment of substantive outcomes, where they were reported.\\
\cmidrule{2-2}
NAPSS \citep{lu-etal-2023-napss} & \textbf{The evidence is current to May 2017.} We included one small study that reported on a limited number of outcomes of interest in this review. ... and inconsistency in how the trial was conducted and reported.  \textbf{The evidence is up to date as of March 2017.} There is insufficient evidence to …\\
\cmidrule{2-2}
LT (Sentence Level) \citep{kang-hashimoto-2020-improved} & This review did not identify any trials that used transabdominal amnioinfusion outside of clinical trials. The evidence in this review can neither support nor refute the use of transcervical amnion infusions for chorioamnionitis and to assess the safety of this intervention or women’s satisfaction.\\
\cmidrule{2-2}
LT (Token Level) & We included one small trial (with data from 34 women) comparing transcervical amnioinfusion with no amnioticinfusion. The trial was considered to be at a high risk of bias overall, inconsistency in the reporting and lack of information on blinding. We did not identify any trials that used transabdominal amniotics (low-quality evidence). We assessed using GRADE (postpartum endometritis, neonatal encephalopathy, admission to intensive/high care) was not reported in the included trial.\\
\cmidrule{2-2}
Drop Sentence &  \textbf{We searched for evidence from randomised controlled trials on 31 May 2017. The evidence in this review is up-to-date at 31 December 2017.} … The overall quality of the evidence was low, with small numbers of women in the trial and a lack of information on blinding.\\
\cmidrule{2-2}
Drop Example & We found one trial of transcervical amnioinfusion, with data from 34 women who matched our question. The trial was considered to be at a high risk of bias overall, due to small numbers, inconsistency in the reporting and lack of information on blinding. Meta-analysis of the evidence was not possible. The majority of trials in this review were of low quality. For these outcomes, we downgraded the evidence to low - with downgrading decisions based on small numbers and a lack of instructions on how to be sure of whether they were in the intervention group. \\
\bottomrule
\end{tabular}
\caption{\label{Table:Example_Output_1} Sample Report from the Cochrane Test Set}
\end{table*}

\section{Examples of Hallucinations}
\label{Appendix:Example_Hallucinations_Eval}

See Table \ref{Table:Example_Hallucinations_Eval} for model outputs containing hallucinated entities. Such examples would count towards the \% reported in hallucination rate (HR).

\begin{table*}[ht]
\centering
\begin{tabular}{p{1.5cm} p{6.5cm} p{6.5cm}}
\toprule
\textbf{Dataset} & \textbf{Source Text} & \textbf{Model Output} \\
\midrule
XSum & The 48-year-old former Arsenal goalkeeper played for the Royals for four years. He was appointed youth academy director in 2000 and has been director of football since 2003. A West Brom statement said: "He played a key role in the Championship club twice winning promotion to the Premier League in 2006 and 2012." & West Bromwich Albion have appointed \textbf{Steve Evans} as their new director of football following his departure from \textbf{Reading}. \\
\cmidrule{1-3}
MedEasi & Some patients have weight loss, rarely enough to become underweight. Anemia, glossitis, angular stomatitis, and aphthous ulcers are usually seen in these patients. & Anemia, glossitis, angular stomatitis, and a number of other \textbf{erythema migrans} spots are usually present in the mouth. \\ 
\cmidrule{1-3}
CNN/ Dailymail & (CNN)A year ago Bloomberg published a story with the following headline: Mike Pence, a Koch Favorite, Mulls 2016 Run for President. The story ticked off... If that holds true in 2016, "winning" is going to feel very strange. & "\textbf{Mike Downey}: Mike Pence's decision to sign a religious freedom law opens the door to discrimination against gays and lesbians. Downey: There's a powerful wing of his party that wants a conservative as its 2016 candidate and this bill was Pence's way of shoring up his street cred." \\
\bottomrule
\end{tabular}
\caption{\label{Table:Example_Hallucinations_Eval} Examples of unsupported entities generated by models are highlighted in \textbf{bold}, which are reflected in the Hallucination Rate (HR) metric}
\end{table*}

\end{document}